\documentclass[11pt]{article}

\usepackage[T1]{fontenc}
\usepackage[utf8]{inputenc}
\usepackage{graphicx}
\usepackage{amsmath}
\usepackage{amssymb}
\usepackage[margin=1in]{geometry}
\usepackage[colorlinks=true, citecolor=blue, urlcolor=blue, linkcolor=blue]{hyperref}

\renewcommand{\cite}[1]{\textsuperscript{#1}}

\title{\textbf{Machines that know they are aging: a framework for hardware-aware autonomous intelligence}}

\author{%
  Cheng Siong Chin\textsuperscript{1,*}, Jianhua Zhang\textsuperscript{2}, Mohan Venkateshkumar\textsuperscript{3}\\[4pt]
  \small\textsuperscript{1}Faculty of Science, Agriculture, and Engineering, Newcastle University Singapore, Singapore 828608\\
  \small\textsuperscript{2}School of Information and Control Engineering, Qingdao University of Technology, Qingdao 266520, China\\
  \small\textsuperscript{3}Department of EEE, Amrita School of Engineering, Amrita Vishwa Vidyapeetham, Chennai, India\\[4pt]
  \small\textsuperscript{*}Corresponding author: \texttt{cheng.chin@newcastle.ac.uk}
}
\date{}

\begin{document}

\maketitle

\begin{abstract}
\noindent\textit{Autonomous systems inevitably age, yet their artificial intelligence typically assumes hardware remains in its original condition. Batteries degrade, sensors drift, processors accumulate timing errors, and memory reliability declines, creating a growing mismatch between assumed and actual capability. This can lead to agnostic collapse, where mission failure arises from accumulated hardware degradation rather than a single component fault. We propose Aging-Aware Autonomous Intelligence (AAAI), a framework that integrates hardware health directly into reasoning, planning, and mission execution. AAAI is built on three pillars: hardware self-awareness, which continuously estimates the health of power, sensing, memory, and computation subsystems using physics-of-failure models; self-adaptive reasoning, which adjusts inference complexity, planning horizon, and task priorities according to remaining hardware capability; and survival-centric intelligence, which allocates remaining operational life across mission objectives through performance optimization, resource conservation, and graceful degradation. Rather than introducing new hardware, AAAI unifies prognostics, lifecycle management, and hardware-aware computing into a closed-loop cognitive architecture. We argue that such integration is essential for autonomous systems operating in inaccessible or safety-critical environments, including space missions, marine robotics, and implantable medical devices. By enabling machines to recognize and respond to their own aging, AAAI improves resilience, extends operational lifetime, and supports safer, more graceful mission completion.}
\end{abstract}

\section*{Introduction}

Consider an autonomous underwater vehicle that has spent two years inspecting subsea pipelines. Its battery retains only a fraction of its original capacity, its depth sensors have drifted, and its processor runs hot from continuous computation. Yet the software remains unaware, still planning missions as though the system were new, still attempting tasks that degraded hardware can no longer reliably perform. This reflects a gap in autonomous systems design. Hardware degradation is well-characterized across battery electrochemistry, sensor drift, and thermal aging in embedded processors, yet many deployed systems do not integrate real-time hardware state into planning and decision-making. The result is a growing mismatch between assumed and actual capability that compounds until it surfaces as mission failure.

\section*{The contract nobody reads}

Modern artificial intelligence systems implicitly assume stable hardware: that sensors faithfully measure the world, that batteries deliver their rated energy, that processors execute without error, and that memory retains what it stores. This assumption holds reasonably well in the short term but becomes increasingly unreliable over longer deployment timescales, and the underlying physics has been well-documented for decades. Lithium-ion batteries undergo irreversible electrochemical change with each charge cycle, depositing microscopic films on electrode surfaces that gradually reduce capacity through well-understood Arrhenius activation kinetics\cite{1}. Silicon processors degrade under sustained voltage stress, slowly eroding transistor timing margins\cite{2}. Memory cells accumulate radiation-induced bit errors at rates that rise with temperature and time\cite{3}. Solar panels on spacecraft lose approximately twenty to forty percent of their output to radiation damage within the first decade of operation\cite{4}. These are not edge cases. They are the expected physical trajectory of deployed hardware, and AI systems that do not account for them are, by design, increasingly misaligned with the substrate on which they depend.

Materials scientists and reliability engineers have documented all of this in extraordinary detail. What remains largely unaddressed is the systematic connection of this knowledge to the intelligence layer of autonomous systems. Few AI frameworks have systematically asked: given that hardware is deteriorating in these specific, measurable ways, how should the intelligence layer reason differently? The machines we build will age. The central engineering question is whether the intelligence layer can be designed to incorporate that knowledge.

\section*{The collapse nobody planned}

We term this failure mode agnostic collapse. It does not manifest as a discrete fault event, but rather as a gradual drift in hardware capability that crosses a performance threshold without triggering conventional diagnostics. Hardware capabilities decline silently until they fall below the thresholds the AI has always assumed, and mission performance degrades without warning. In simulations of a marine autonomous vehicle, the pattern is consistent: a hardware-agnostic system maintains strong task performance early in a long mission, then deteriorates steadily as battery capacity and sensor accuracy each decline through ordinary wear. The hardware did not fail. It aged past the point where a hardware-agnostic control system could continue to operate effectively.

Two failure modes produce this outcome. In the first, the system continues attempting computationally demanding operations, including deep-learning inference, long-range planning, and high-resolution sensing, that its diminished hardware can no longer execute reliably. In the second, it overcorrects, defaulting to conservative safe mode well before necessary and squandering usable residual capacity. Both stem from the same underlying absence: the intelligence layer has no mechanism for reasoning about its own physical condition.

Existing approaches offer only partial remedies. Prognostics tools can forecast component end-of-life\cite{5} but knowing that battery capacity is nearly exhausted does not inform how the system should reason or plan in the interim. Hardware redundancy can mask individual component failures\cite{6}, but it is finite and costly to implement on constrained platforms. What has been missing is a cognitive response to physical ageing: an intelligence layer that modifies its reasoning and planning as its hardware changes.

\section*{What existing approaches offer and where they stop}

Recent work in AI-based device health monitoring addresses aspects of this problem but remains at the hardware management level, sensing degradation, adjusting resource allocation, and protecting components from further wear. Cognitive adaptation, in which the system actively reasons about its physical condition and modifies planning and prioritisation accordingly, remains absent. The following maps existing research to the pillar it most closely addresses, making explicit where prior work ends and where AAAI begins.

\subsection*{Pillar 1: Hardware Self-Awareness}

Physics-of-failure modelling provides the theoretical foundation\cite{7}. On-chip sensors now detect early-stage Time-Dependent Dielectric Breakdown, and commercial AI platforms incorporate adaptive agents capable of on-device learning in response to measured hardware state\cite{8}. Negative Bias Temperature Instability tracking provides a real-time index of transistor ageing in production hardware\cite{9,10}. What these contributions do not address is the step from monitoring to thinking: health signals reach engineers or trim local settings, but they do not change how the system reasons, what it chooses to do, or how it distributes its remaining capacity. Bridging that gap is the first function of AAAI.

\subsection*{Pillar 2: Autonomous Lifecycle Management}

Existing algorithms autonomously balance active and idle hardware components to manage performance against longevity\cite{11,12}. These approaches are technically feasible in deployed systems but operate at the resource-allocation layer without connecting hardware state to higher-level reasoning or mission planning.

\subsection*{Pillar 3: Proactive Life Extension}

ReaLM\cite{13} identifies operating points at which model performance is preserved despite reduced hardware power, exploiting neural network tolerance to numerical perturbations as a functional buffer. Ageing-aware voltage scaling applies this logic at the circuit level, slowing transistor wear without triggering functional failure\cite{9}. Each addresses a distinct layer without integrating across them.

\subsection*{Pillar 4: Cross-Layer Co-Design}

HAWQ-V2\cite{14} performs layer-wise sensitivity analysis, allocating higher numerical precision to quantisation-vulnerable layers while compressing more resilient ones. Hardware co-design workflows iterate between software and hardware specifications until a jointly feasible optimum is reached. Sensitivity-guided task routing directs demanding operations toward chip regions with the greatest remaining reliability margin\cite{15}. What these approaches collectively lack is runtime coupling between hardware health and system reasoning. Co-design fixes the relationship between software and hardware at build time, but does not equip a deployed system to think or plan differently as its physical capabilities diminish.

\section*{The cognitive gap}

Existing research addresses hardware health monitoring and model-complexity matching but stops short of cognitive integration: a unified architecture in which continuous multi-subsystem health estimation actively shapes reasoning, planning, and mission strategy in real time. To the best of our knowledge, a system that integrates all four layers and explicitly links hardware health decisions to higher-level mission objectives has not been reported. The contribution of AAAI lies in the architectural design that connects these existing components into a coherent, closed-loop cognitive system.

\section*{Three ways to grow old gracefully}

\begin{figure}[htbp]
  \centering
  \includegraphics[width=0.85\textwidth]{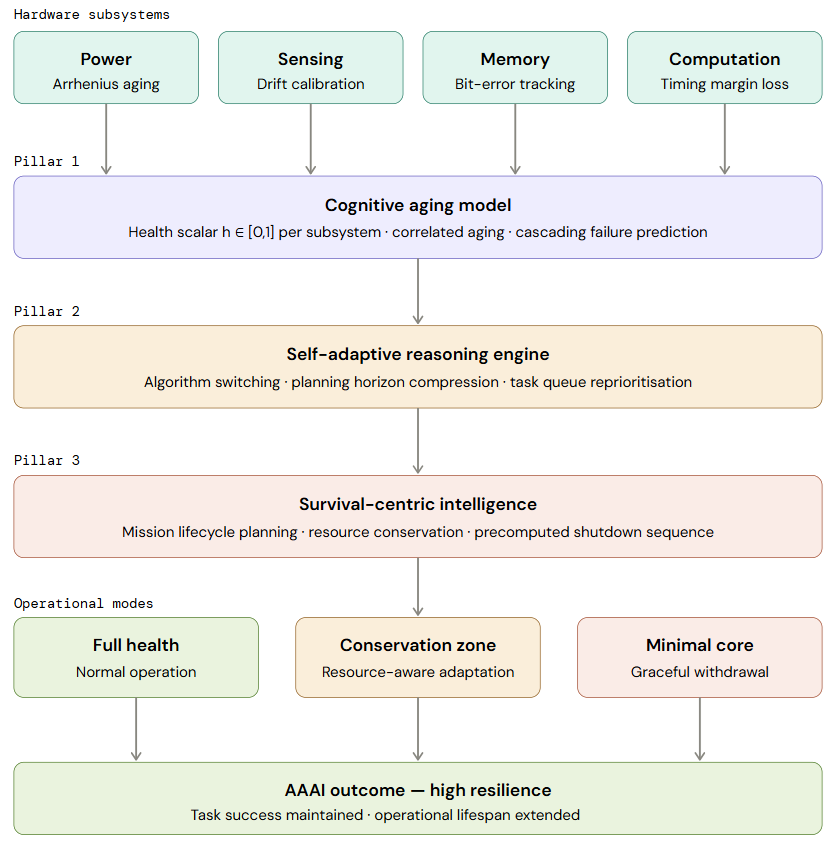}
  \caption{The Aging-Aware Autonomous Intelligence (AAAI) Architecture. The framework comprises three functional pillars that connect physical hardware state to cognitive behaviour. Pillar 1, the cognitive aging model, represents the condition of each subsystem, namely power, sensing, memory, and computation, as a continuous health scalar $h \in [0, 1]$, updated in real time using physics-of-failure models. Pillar 2, the self-adaptive reasoning engine, adjusts inference complexity, planning horizon, and task prioritisation in proportion to current hardware capability. Pillar 3, the survival-centric intelligence layer, distributes remaining hardware life across mission objectives through three operational modes: full-health operation, resource-conservation, and graceful withdrawal to a minimal functional core. Together, the three pillars maintain task performance as hardware degrades, in contrast to the aging-agnostic baseline, which undergoes abrupt mission failure as accumulated degradation silently crosses an unmonitored threshold, the condition we term agnostic collapse.}
  \label{fig:architecture}
\end{figure}

The first pillar of Aging-Aware Autonomous Intelligence (AAAI) is genuine hardware self-knowledge (Fig.~\ref{fig:architecture}). We represent the condition of each critical subsystem, namely: power, sensing, memory, and computation, as a number between one (fully healthy) and zero (failed), continuously updated using the same physics-of-failure models that materials scientists apply to predict degradation\cite{7}. AAAI integrates and extends the four strands of prior work surveyed above into a unified cognitive architecture that connects physical hardware state to adaptive reasoning, task prioritisation, and mission planning.

Health indicators across subsystems do not evolve independently. Environmental stresses frequently affect multiple components simultaneously: elevated temperatures accelerate battery degradation while reducing processor reliability, and salt exposure damages sensors and surrounding circuitry alike. Monitoring these coupled degradation dynamics as an interconnected process, rather than tracking components in isolation, enables earlier identification of emerging platform-wide failures before any individual component reaches a critical threshold.

\section*{Think within your limits}

Hardware self-awareness confers limited autonomous operational benefit unless it actively informs system behaviour. The second pillar of AAAI provides a reasoning layer that dynamically modulates task scope and complexity as a function of available hardware capacity. Under nominal computational and memory conditions, AAAI deploys high-capacity deep learning models with extended planning horizons. As measured hardware capacity declines, the system responds proportionally: substituting lightweight inference architectures, contracting its planning horizon, and reordering its task queue to prioritise operations with the highest utility-to-cost ratio. This adaptation is not implemented as a discrete mode transition but as a continuous, proportional control response that tracks the degrading hardware envelope in real time\cite{16}. At degradation thresholds where a hardware-agnostic system would undergo functional collapse, AAAI preserves operational continuity by dynamically aligning computational demand with remaining physical capacity.

\section*{Survive with purpose}

The third pillar extends the mission objective to include active management of remaining operational capacity. Rather than treating operational lifespan as a passive consequence of hardware endurance, AAAI treats it as a resource to be explicitly allocated across competing mission priorities, with the objective of maximising mission utility over the remaining deployment period. This framework translates into three adaptive operational states. Under healthy conditions, the system operates normally with full capability. As degradation emerges, it transitions into a conservation phase that reduces resource consumption while preserving mission-critical functions. In the final stage, the system progressively shuts down non-essential subsystems according to a precomputed prioritisation strategy, restricts behaviour to a minimal operational core, and reduces communication frequency to maximise remaining operational duration. This is graceful degradation in the most literal sense: not a collapse, but a planned withdrawal that preserves useful function to the very last moment\cite{17}.

\section*{What this means in practice}

In low Earth orbit, microsatellites are continuously exposed to thermal cycling and ionising radiation, both of which progressively degrade processors, memory, and power subsystems. An ageing-aware architecture can adapt its operational profile in response, for example by reducing processor activity during thermally stressful orbital phases, adjustments that may cumulatively extend functional lifetime over longer deployment periods. Comparable challenges arise in offshore robotics and autonomous marine systems, where corrosion, hydrostatic pressure, and mechanical vibration gradually compromise structural and electronic integrity. Conventional systems typically detect failure only after performance degradation has become critical. An intelligence layer that continuously models component health can instead proactively reallocate resources, reduce operational strain, and preserve essential functionality under deteriorating conditions.

This capability is particularly consequential in implantable medical devices such as cardiac pacemakers, where premature battery or component failure may require surgical intervention rather than routine maintenance. Ageing-awareness therefore represents not only an engineering strategy for extending operational endurance, but also a mechanism for improving safety and reliability in life-critical applications.

What unites these domains is a shared vulnerability to a common oversight: the assumption that capability remains constant until the moment it fails. Addressing this requires no new materials, faster processors, or additional power sources. It requires primarily that the intelligence layer act on what materials scientists have long established: hardware ages predictably, and that predictability should drive behaviour.

Ageing-awareness also introduces a category of risk that must be acknowledged alongside its potential benefits. Autonomous systems empowered to adjust their own decision-making in response to inferred hardware state may, without sufficient constraints, make consequential operational choices that were neither anticipated nor authorised by their designers. This concern is not theoretical. Incidents involving commercial automation systems, including mode-confusion events in aviation and the limitations exposed in early autonomous vehicle deployments, demonstrate how small discrepancies between assumed and actual operating conditions can propagate into safety-critical outcomes. In the context of AAAI, transitions between operational modes, particularly the shift to graceful withdrawal, must be governed by explicit, auditable criteria and remain subject to human oversight wherever possible. The framework is not intended to replace human judgement in high-stakes decisions, but to inform it by surfacing hardware state information that would otherwise be unavailable to operators. A rigorous safety architecture, encompassing formal specification of mode-transition conditions, fail-safe defaults, and human-in-the-loop confirmation thresholds, is an essential precondition for deployment in safety-critical contexts such as medical devices or crewed spacecraft.

\section*{Conclusion}

Autonomous systems are increasingly deployed in environments where maintenance or replacement is difficult or impossible, yet most contemporary AI architectures treat computation as independent of the physical hardware on which it runs. This assumption becomes increasingly unreliable as systems enter safety-critical and long-duration settings ranging from orbital infrastructure and subsea robotics to implantable medical devices. The central argument of this Comment is that intelligence can no longer be separated from the material condition of the platform that sustains it. Hardware degradation is not an exceptional event, but a continuous physical process governed by well-established principles of materials ageing, electrochemistry, thermal stress, and radiation damage. The scientific capacity to characterise and predict these processes exists. What is largely absent is an AI framework capable of integrating this knowledge into operational decision-making in real time. Ageing-aware artificial intelligence addresses this gap by continuously monitoring component health, modelling correlated degradation across subsystems, and adapting behaviour in response to declining physical capacity. Systems designed on this basis may extend operational lifetime, improve resilience, and reduce the probability of unplanned mission failure. More broadly, they represent a shift in how autonomy is conceived: not as computation independent of its physical substrate, but as embodied intelligence that is constrained by, and responsive to, the irreversible realities of physical ageing. Several critical limitations remain. The accuracy of physics-of-failure models varies with deployment conditions, coupled degradation effects are difficult to characterise exhaustively in advance, and the computational overhead of continuous health monitoring must itself be accounted for against declining hardware capacity.

Developing these capabilities will require close integration across materials science, reliability engineering, control theory, embedded systems, and artificial intelligence. Recent advances in in situ AI-based device health monitoring indicate that elements of this integration are already emerging; the contribution of AAAI lies in unifying these strands within a coherent cognitive framework that links physical state to adaptive reasoning. Establishing that interdisciplinary foundation may prove to be one of the defining challenges for the next generation of autonomous technologies.


\end{document}